\documentclass{article}

\usepackage{microtype}
\usepackage{graphicx}
\usepackage{subfigure}
\usepackage{booktabs} 
\usepackage{textcomp}
\usepackage{comment}
\usepackage{dblfloatfix}    

\usepackage{hyperref}

\usepackage[accepted]{icml2020_v2}
\usepackage{amsmath}
\usepackage{lineno}

\DeclareMathOperator*{\argmin}{arg\,min}

\icmltitlerunning{Robust binary classification with the 01 loss}

\begin{document}

\twocolumn[
\icmltitle{Robust binary classification with the 01 loss}



\icmlsetsymbol{equal}{*}

\begin{icmlauthorlist}
\icmlauthor{Yunzhe Xue}{to}
\icmlauthor{Meiyan Xie}{to}
\icmlauthor{Usman Roshan}{to}
\end{icmlauthorlist}

\icmlaffiliation{to}{Department of Computer Science, New Jersey Institute of Technology, Newark, NJ 07102, USA}

\icmlcorrespondingauthor{Usman Roshan}{usman@njit.edu}


\vskip 0.3in
]



\printAffiliationsAndNotice{}  

\begin{abstract}
The 01 loss is robust to outliers and tolerant to noisy data compared to convex loss functions. We conjecture that the 01 loss may also be more robust to adversarial attacks. To study this empirically we have developed a stochastic coordinate descent algorithm for a linear 01 loss classifier and a single hidden layer 01 loss neural network. Due to the absence of the gradient we iteratively update coordinates on random subsets of the data for fixed epochs. We show our algorithms to be fast and comparable in accuracy to the linear support vector machine and logistic loss single hidden layer network for binary classification on several image benchmarks, thus establishing that our method is on-par in test accuracy with convex losses. We then subject them to accurately trained substitute model black box attacks on the same image benchmarks and find them to be more robust than convex counterparts. On CIFAR10 binary classification task between classes 0 and 1 with adversarial perturbation of 0.0625 we see that the MLP01 network loses 27\% in accuracy whereas the MLP-logistic counterpart loses 83\%. Similarly on STL10 and ImageNet binary classification between classes 0 and 1 the MLP01 network loses 21\% and 20\% while MLP-logistic loses 67\% and 45\% respectively. On MNIST that is a well-separable dataset we find MLP01 comparable to MLP-logistic and show under simulation how and why our 01 loss solver is less robust there. We then propose adversarial training for our linear 01 loss solver that significantly improves its robustness on MNIST and all other datasets and retains clean test accuracy. Finally we show practical applications of our method to deter traffic sign and facial recognition adversarial attacks. We discuss attacks with 01 loss, substitute model accuracy, and several future avenues like multiclass, 01 loss convolutions, and further adversarial training. 
\end{abstract}

\section{Introduction}
Adversarial attacks present a  challenge to machine learning algorithms typically based on convex losses. State of the art classifiers like the support vector machine \cite{cortes95} and neural networks \cite{krizhevsky2012imagenet} achieve high accuracies on test data but are also vulnerable to adversarial attacks based on minor perturbations in the data \cite{goodfellow2014explaining,papernot2016limitations,kurakin2016adversarial,carlini2017towards,brendel2017decision}. To counter adversarial attacks many defense methods been proposed with adversarial training being the most popular \cite{szegedy2013intriguing}. This is known to improve model robustness but also tends to lower accuracy on clean test data that has no perturbations \cite{raghunathan2019adversarial,zhang2019theoretically,raghunathan2019adversarial}. 


The robustness of outliers to the 01 loss is well known \cite{bartlett04}. Convex loss functions such as least squares are affected by both correct and incorrectly classified outliers and hinge is affected by incorrectly classified outliers whereas the 01 loss is robust to both \cite{xie2019,icml13optimize}. In addition to being robust to outliers the 01 loss is also robust to noise in the training data \cite{manwani2013noise,ghosh2015making} and under this loss minimizing the empirical risk amounts to minimizing the empirical adversarial risk \cite{lyu2019curriculum,hu2016does} with certain assumptions of noise. We conjecture that these properties may translate to robustness against black box adversarial attacks that typically succeed in fooling state of the art classifiers \cite{papernot2017practical}.

To test this we first develop stochastic coordinate descent solvers for 01 loss based upon prior work \cite{xie2019}. 
We also extend the previous work to a non-linear single hidden layer 01 loss network that we call MLP01. For the task of binary classification on standard image recognition benchmarks we show that our linear 01 loss solver and the MLP01 loss are both as accurate as their convex counterparts, namely the linear support vector machine and the logistic loss single hidden layer network. We then subject all methods to a substitute model black box attack \cite{papernot2017practical} and find both our 01 loss models (linear and non-linear) to be more robust than hinge and logistic. 

We find that on separable image datasets like MNIST our model offers little advantage and demonstrate under simulation why this happens. We then conduct adversarial training of our linear model and show it increases its robustness on MNIST and all other datasets while retaining clean test accuracy. We also show applications to deter street sign and facial recognition adversarial attacks. We describe below our methods followed by results and discussion.

\section{Methods}

\subsection{Background}
The problem of determining the hyperplane with minimum number of misclassifications
in a binary classification problem is known to be NP-hard \cite{ben03}.
In mainstream machine learning literature this is called minimizing the 01 loss
\cite{kernel01} as given in Objective~\ref{obj1},

\begin{equation}
\frac{1}{2n}\argmin_{w,w_0} \sum_i (1-sign(y_i(w^Tx_i+w_0)))
\label{obj1}
\end{equation}

where $w \in R^d$, $w_0 \in R$ is our hyperplane solution, and $x_i \in R^d, y_i\in \{+1,-1\}.\forall i=0...n-1$ are our training data. Popular linear classifiers such as the linear support 
vector machine, perceptron, and logistic regression \cite{alpaydin} can be considered 
as convex approximations to this problem that yield fast gradient descent solutions \cite{bartlett04}. 
However, they are also more sensitive to outliers than the 01 loss \cite{bartlett04,icml13optimize,xie2019}. 

We extend the 01 loss to a simple single hidden layer neural network with $k$ hidden nodes and sign activation that we call the MLP01 loss. This objective can be given as

\begin{equation}
\small
\frac{1}{2n}\argmin_{W, W_0, u,u_0} \sum_i (1-sign(y_i(u^T(sign(W^Tx_i+W_0))+u_0)))
\label{obj2}
\end{equation}

where $W \in R^{d\times k}$, $W_0 \in R^k$ are the hidden layer parameters, $u\in R^k,u_0\in R$ are the final layer node parameters, $x_i \in R^d, y_i\in \{+1,-1\}.\forall i=0...n-1$ are our training data, and $sign(v\in R^k)=(sign(v_0), sign(v_1),...,sign(v_{k-1}))$. 

We solve both problems with stochastic coordinate descent based upon earlier work \cite{xie2019}. 

\paragraph{Other work on 01 loss solvers}
Aside from the stochastic coordinate descent \cite{xie2019} that we build upon other attempts have been made to optimize the 0/1 loss. These include boosting \cite{nips01optimize}, integer programming \cite{mixedint01}, an approximation algorithm \cite{approx01}, a random coordinate descent method \cite{ijcnn01optimize}, and a branch and bound method that is the most recent from 2013 \cite{icml13optimize}. The above previous works cover various strategies to solve 01 loss but lack on-par test accuracy with convex solvers on real data. We obtained a Matlab implementation of the branch and bound method and found it to be slow - it did not finish after several hours of runtime, as also cautioned by authors in their code. The random coordinate descent code \cite{ijcnn01optimize} requires GNU C compiler (gcc) version 3.0 to compile whereas current supported versions are above 4.0. 

\subsection{Stochastic coordinate descent (SCD01)}
We use the stochastic coordinate descent for 01 loss \cite{xie2019} called SCD01 to drive our linear and non-linear 01 loss solvers. In the Supplementary Material we fully describe the algorithm for reference including the optimal threshold algorithm. Briefly, we iteratively randomly select a subset of the training data in each epoch and run coordinate descent in each iteration. Our coordinate descent shown in Algorithm~\ref{ls} differs from previous work in how we update the coordinates. In previous work \cite{xie2019} authors update each coordinate until there is no change in the objective. We randomly update a pool of coordinates by one step and pick the one with the greatest decrease in the objective (with ties decided randomly). 

\begin{algorithm}[!h]
\small
\caption{\small Coordinate descent} 
\label{ls}
\textbf{Input: } Data (feature vectors) $x_i \in R^d$ for $i=0..n-1$ 
with labels $y_i \in \{+1,-1\}$, $w_{inc} \in R$ (set to 0.17 by default), size of pooled features to update $k$ (set to 128 by default), vector $w \in R^d$ and $w_0 \in R$\\
\textbf{Output: } Vector $w \in R^d$ and $w_0 \in R$ \\
\textbf{Procedure: }
\begin{algorithmic}
\STATE 1. Initialization: If $w$ is null then let each feature $w_i$ of $w$ be uniformly drawn from $(-1,1)$. 
We set $\|w\|=1$ and throughout our search ensure that $\|w\|=1$ by renormalizing each time $w$ changes.
\STATE 2. Let the number of misclassified points with negative $w^Tx_i$ be $errorminus=0$
and those with positive $w^Tx_i$ be $errorplus=0$.  These are later used in the Optimal Threshold algorithm called \emph{Opt} (see Supplementary Material) for fast update of our objective. 
\STATE 3. Compute the initial data projection $w^Tx_i, \forall i=0..n-1$,
sort the projection with insertion sort, and initialize $(w_0,obj)=Opt(w^Tx, y, 0,n-1)$. We also record the value of $j$ for the optimal $w_0=(w^Tx_j+w^Tx_{j+1})/2$.
\STATE 4. Set $prevobj = \infty$.
\WHILE {$prevobj - obj > 0$} 
	\STATE Set $prevobj=obj$
	\STATE Randomly pick $k$ of the $d$ feature indices.
	\FORALL{selected features $w_i$ we update them}
		\STATE 1. Assume the optimal $w_0=(w^Tx_j+w^Tx_{j+1})/2$
		\STATE 2. Set $start=w^Tx_{j-10}$ and $end=w^Tx_{j+10}$ 
		\STATE 3. Modify coordinate $w_i$ by $w_{inc}$, compute data projection $w^Tx_i \forall i=0..n-1$, and 
		sort the projection with insertion sort
		\STATE 4. Set $(w_0,obj)=Opt(w^Tx, y, start,end)$ and record this value for feature $w_i$
		\STATE 5. Reset $w_0$ to try the next coordinate
	\ENDFOR
	\STATE Pick the coordinate whose update gives the largest decrease in the objective and set $(w_0,obj)$ to the values given by the best coordinate with ties decided randomly.
\ENDWHILE
\end{algorithmic}
\end{algorithm}

\subsection{Single hidden layer 01 loss network (MLP01)}
We extend the stochastic coordinate descent solver to a single hidden layer network with $k$ hidden nodes that we call MLP01 (see Algorithm~\ref{mlp01}). For each random batch of the training data we train the final node followed by each hidden node using our Coordinate Descent algorithm above (Algorithm~\ref{ls}). We set 20 hidden nodes ($h=20$) in our experiments.

\begin{algorithm}[!b]
\small
\caption{\small Stochastic coordinate descent for single hidden layer 01 loss network} 
\label{mlp01}
\textbf{Input: } Data (feature vectors) $x_i \in R^d$ with labels $y_i \in \{+1,-1\}$, number of hidden nodes $h$ (set to 20 by default), number of votes $rr \in N$ (Natural numbers), number of iterations per vote $it \in N$ (set to 2000 by default), batch size as a percent of training data $p \in [0,1]$ (set to 0.75 by default) , $w_{inc} \in R$ (set to 0.1 by default) and $w_{inc2} \in R$ (set to 0.02) \\
\textbf{Output: } Total of $rr$ sets of $(bestW\in R^{k\times d}, bestW_0\in R^k, bestu\in R^k, bestu_0\in R)$ after each vote \\
\textbf{Procedure: }
\begin{algorithmic}
\STATE 1. Initialize all network weights $W,u$ to random values from the uniform distribution $(-1,1)$.
\STATE 2. Set network thresholds $W_0$ and $u_0$ to the median projection value on their corresponding weight vectors.
\WHILE {$j < rr$} 
	\STATE  Set $bestW=null, bestW_0=null, bestw=null, bestw_0=null, bestloss=\infty$
	\FOR{$i = 0$ to $it$}
		\STATE Randomly pick $p$ percent of rows as input training data.
		\STATE Run the Coordinate Descent Algorithm~\ref{ls} for the final output node $u$ to completion starting with the values of $u$ and $u_{0}$ from the previous call to it (if $i==0$ we set $u=null$). 
		\FOR{$k=0$ to $h$}
			\STATE Run the Coordinate Descent Algorithm~\ref{ls} to completion starting with the values of $w_k$ ($k^{th}$ column in $W$) and $w_{k0}$ ($k^{th}$ entry in $W_0$) from the previous call to it (if $i==0$ we set $w_k=null$).
		\ENDFOR
		\STATE Calculate Objective~\ref{obj2} on the full input training set 
		\IF {$objective(W,W_0,u,u_0) < objective(bestW,bestW_0, bestu, bestu_0)$}
			\STATE Set $bestW=W$, $bestW_0=W_0$, $bestu=u$, $bestu_0=u_0$, and $bestloss=objective(bestW,bestW_0, bestu, bestu_0)$
		\ENDIF
	\ENDFOR
	\STATE Output ($bestW$, $bestW_0$, $bestu$, $bestu_0$) 
	\STATE Set $j=j+1$.
\ENDWHILE
\STATE We output all sets of $(bestW, bestW_0, bestu, bestu_0)$ across the votes. We can use the first set or the majority vote of all sets for predictions. 
\end{algorithmic}
\end{algorithm}

\subsection{Majority vote 01 loss}
Due to the non-uniqueness of 01 loss and randomness of our solvers both our methods will return different solutions when initialized with different seeds for the random number generator. Thus we take the majority vote of multiple runs which we see as inherently necessary due to the nature of 01 loss. We call our methods SCD01majvote that has 100 votes and MLPmajvote that has 32 votes (input $rr=32$ in Algorithm~\ref{mlp01}). 

\subsection{Adversarial training SCD01}
We apply the basic iterative adversarial training described earlier \cite{kurakin2016adversarial} to our SCD01 algorithm. The adversarial training objective is actually a min-max objective: we minimize the empirical risk across the maximum distortion of the input data. The iterative training that has been proposed earlier and used by us below is a heuristic to the min-max problem.

\subsection{Black box adversarial attack}
Since the 01 loss model has no gradient we cannot use white box gradient based attacks. Instead we resort to a black box strategy that uses the gradient of a substitute model to generate adversaries \cite{papernot2017practical}. In this setting we start with a small subset of the test data (200 samples) from which we iteratively learn the substitute model parameters. In each iteration (epoch) we generate adversaries from the remaining test and attack the target model. This strategy can be effective as long as the substitute model is at least as accurate on the test data as the target model. This indicates that the substitute model is accurately modeling the target model at least on the samples it is trained upon and so its gradient is likely to produce effective adversaries. 

For the substitute model we use a two hidden layer neural network each with 200 nodes per layer as in previous work \cite{galloway2017attacking}. We implement this using the multilayer perceptron class in the Python scikit-learn toolkit \cite{scikit}. In the Supplementary Material we fully describe our black box attack model. We set the distortion in the adversarial images to $\epsilon=0.0625$ for CIFAR10, STL10, ImageNet and $\epsilon=0.3$ for MNIST. In the Supplementary Material we provide results for lower values of $\epsilon$ and see similar results as here.
\vspace{-.05in}

\section{Results}
We study all methods for the task of binary classification of classes 0 and 1 on four popular image classification benchmarks. We compare our 01 loss methods to their convex counterparts on each benchmark. We run the LinearSVC algorithm (SVM) in Python scikit-learn toolkit \cite{scikit} with cross-validated $C$. We use the Python scikit-learn multilayer perceptron with logistic loss (MLP) and one hidden layer of 20 the same as the MLP01 network. We train our MLP with stochastic gradient descent and learning parameters set to optimize the cross-validation accuracy. We also study a majority vote SVM and MLP by running them a 100 times on bootstrapped samples (bagging) and find no improvement in robustness compared to the single runs. We obtained the previous stochastic coordinate descent 01 loss solver \cite{xie2019} that we compare to our new one. All of our code and data are freely available from our GitHub site \url{https://github.com/zero-one-loss/01loss}.

\vspace{-.07in}
\begin{algorithm}
\caption{\small Iterative adversarial training for SCD01}
\small
\label{adv}
\textbf{Input: } Model vector $w \in R^d$ and $w_0 \in R$ from SCD01, data (feature vectors) $x_i \in R^d$ with labels $y_i \in \{+1,-1\}$ for $i=0..n-1$\\
\textbf{Output: } 100 adversarially trained models $(w',w'_0)$\\
\textbf{Procedure: }
\begin{algorithmic}
\FOR{each of 100 iterations}
	\STATE Randomly select 10\% of the input training examples and produce adversaries for each selected datapoint
	\FOR{each datapoint $x_i$ }
		\STATE Obtain prediction of $x_i$ as $y'_i=w^Tx_i + w_0$
		\STATE Create adversary $x_i' = x_i + (-y'_i)\frac{w}{||w||}$
	\ENDFOR
	\STATE Run SCD01 on the clean plus adversarial samples selected above starting the search from $w$ in the previous iteration.
\ENDFOR
\STATE We now have 100 SCD01 models on the clean ($x_i$) plus adversarial ($x'_i$) examples 
\STATE \textbf{return} (all 100 ($w',w'_0$) for majority vote output)
\end{algorithmic}
\end{algorithm}
\vspace{-.2in}



\subsection{CIFAR10, STL10, and ImageNet}
\begin{figure}[h!]
\begin{center}
\centerline{\includegraphics[width=\columnwidth]{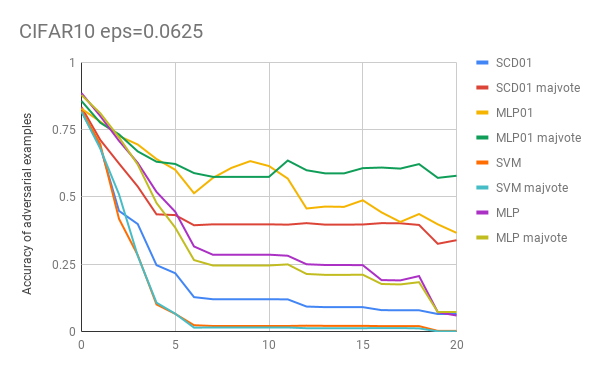}}
{\footnotesize
\begin{tabular}{ccccc}
Epoch & SCD01majvote & SVM & MLP01majvote  & MLP \\
0 & .83 & .82 & .86 & .89 \\
20 & .34 & 0 & .59 & .06 \\
\end{tabular}
}
\centerline{\includegraphics[width=\columnwidth]{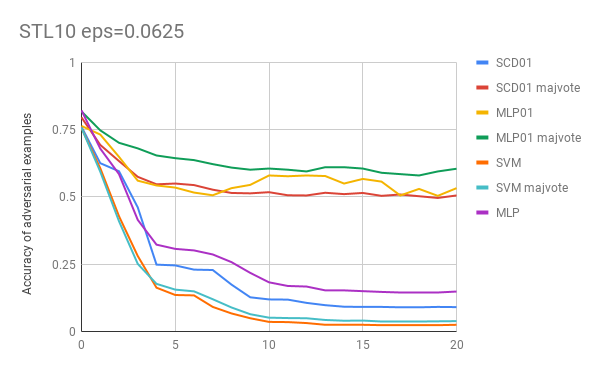}}
{\footnotesize
\begin{tabular}{ccccc}
Epoch & SCD01majvote & SVM & MLP01majvote  & MLP \\
0 & .8 & .76 & .82 & .82 \\
20 & .51 & .03 & .61 & .15 \\
\end{tabular}
}
\centerline{\includegraphics[width=\columnwidth]{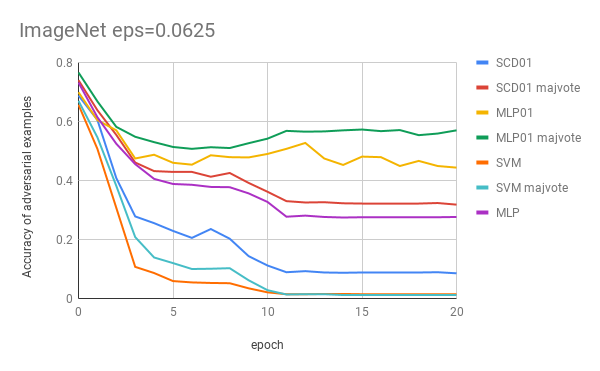}}
{\footnotesize
\begin{tabular}{ccccc}
Epoch & SCD01majvote & SVM & MLP01majvote  & MLP \\
0 & .74 & .66 & .77 & .73 \\
20 & .32 & .02 & .57 & .28 \\
\end{tabular}
}
\vspace{-.15in}
\caption{Accuracy of adversarial samples generated at each epoch during substitute model training on CIFAR10, STL10, and ImageNet. At epoch 0 we have the accuracy of the target model on clean test data (without adversaries) as shown in the tables.}
\label{alldata}
\end{center}
\vskip -0.2in
\end{figure}

We start with results for binary classification of classes 0 and 1 in CIFAR10, STL10, and ImageNet. Between classes 0 and 1 we have in CIFAR10 \cite{krizhevsky2009learning} we have 10,000 $32\times32\times3$ training images and 2000 test ones and in STL10  \cite{coates2011analysis} we have 1000 $96\times96\times3$ training images and 1600 test. In ImageNet classes 0 and 1 contain about 2580 $256\times256\times3$ training images and 100 test ones. We change the split so as to increase the test data size so that we can better train the black box attack substitute model. We divide the training set into two parts: the first containing 1280 for training and 1300 for test.  

\begin{figure}[h!]
\begin{center}
\begin{tabular}{cccc}
Clean & MLP01 majvote & MLP & SVM \\
\includegraphics[scale=1.5]{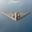} & \includegraphics[scale=1.5]{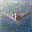} & \includegraphics[scale=1.5]{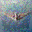} & \includegraphics[scale=1.5]{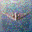} \\
\includegraphics[scale=.55]{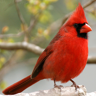} & \includegraphics[scale=.55]{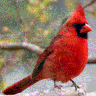} & \includegraphics[scale=.55]{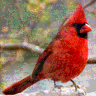} & \includegraphics[scale=.55]{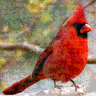} \\
\includegraphics[scale=.25]{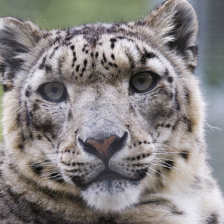} & \includegraphics[scale=.25]{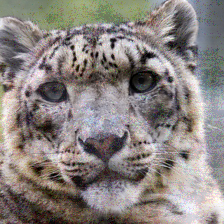} & \includegraphics[scale=.25]{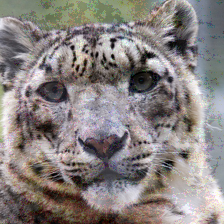} & \includegraphics[scale=.25]{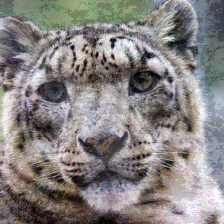} \\
\end{tabular}
\vspace{-.1in}
\caption{Clean images from CIFAR10 (top row), STL10 (mid row), and ImageNet (bottom row) and adversarial images obtained by attacking MLP01, MLP, and SVM with $\epsilon=0.0625$. The adversarial images shown here fool SVM and MLP but are correctly classified by MLP01. 
}
\label{sampleimages}
\end{center}
\vskip -0.2in
\end{figure}

In Figure~\ref{alldata} we see that both our linear SCD01 and non-linear MLP01 models have comparable accuracy to the linear SVM and non-linear MLP but are much more robust. In Figure~\ref{sampleimages} we show clean and adversarial images generated by attacking MLP01, MLP, and SVM. We show adversarial images that are correctly classified by MLP01 and wrongly by MLP and SVM. 

\setlength{\tabcolsep}{1pt}


\subsection{MNIST and simulation}

\setlength{\tabcolsep}{6pt}
\begin{figure}[h!]
\begin{center}
\centerline{\includegraphics[width=\columnwidth]{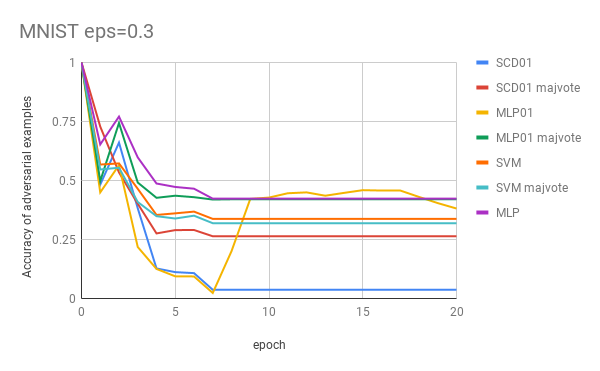}}
{\footnotesize
\begin{tabular}{ccccc}
Epoch & SCD01majvote & SVM & MLP01majvote  & MLP \\
0 & 1 & 1 & 1 & 1 \\
20 & .26 & .34 & .42 & .43 \\
\end{tabular}
}
\vspace{-.1in}
\caption{Accuracy of adversarial samples on MNIST (see Figure~\ref{alldata} caption for more)}
\label{mnist}
\end{center}
\vskip -0.2in
\end{figure}

While the above benchmarks are focused on image classification of arbitrary objects the MNIST benchmark focuses on digit classification and is easier in comparison. Its test accuracy is typically above 99\% for most classifiers. Between classes 0 and 1 (also digits 0 and 1) we have 12,665 training images and 2115 test images each of size $28\times28$. In Figure~\ref{mnist} we see that SCD01 is not as robust as SVM and MLP01 is the same as MLP. We conjecture that this may be due to non-uniqueness of 01 loss on easily separable classes like we see in MNIST. To understand this better we turn to simulated data.

\begin{figure}[h!]
\begin{center}
\begin{tabular}{cc}
\includegraphics[scale=.28]{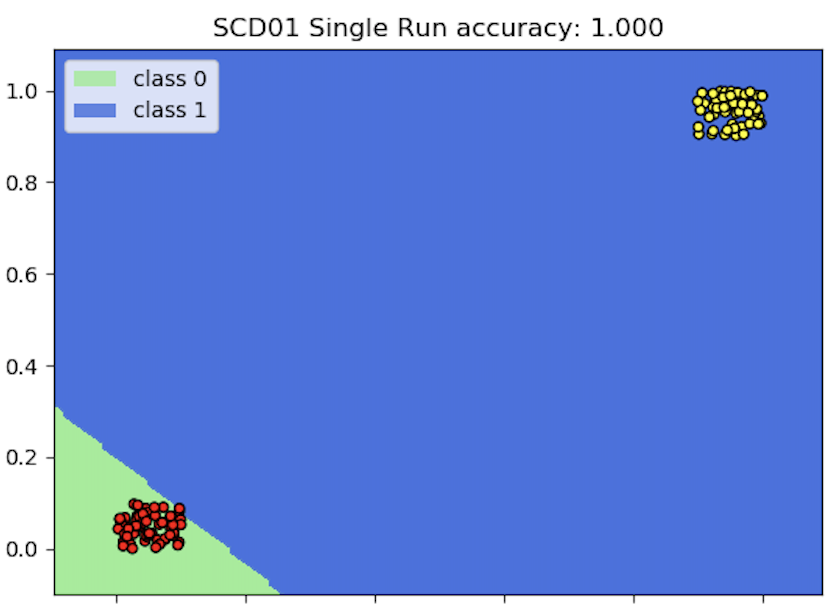} & \includegraphics[scale=.27]{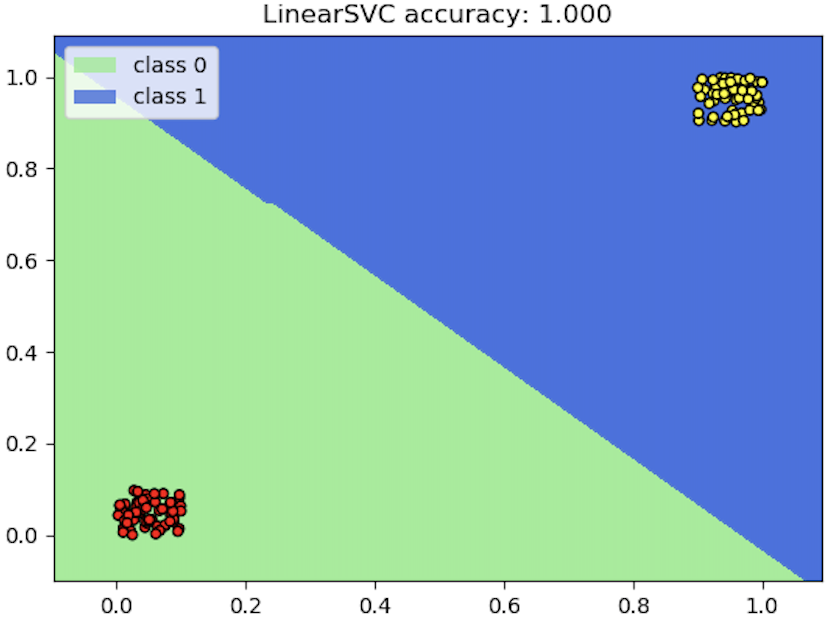} \\
(a) SCD01 simple & (b) SVM simple  \\
\includegraphics[scale=.28]{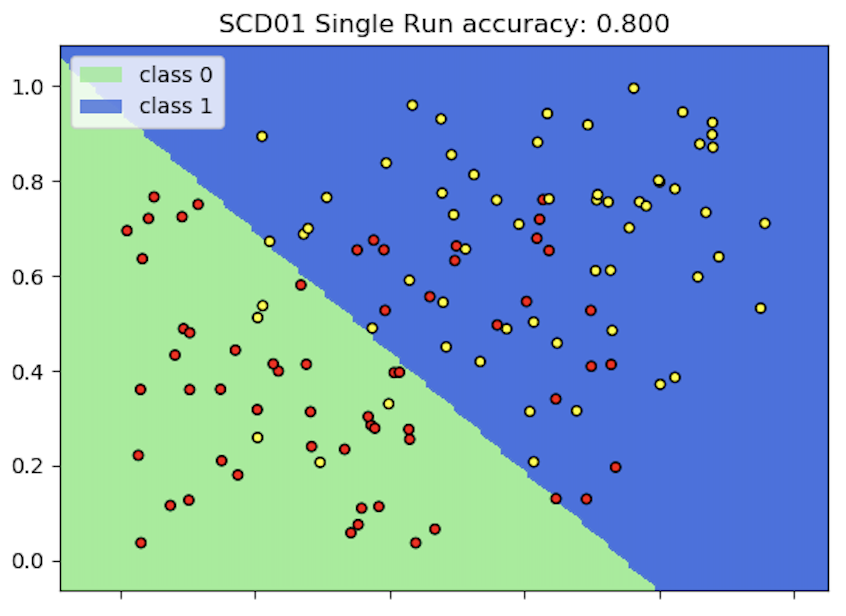} & \includegraphics[scale=.27]{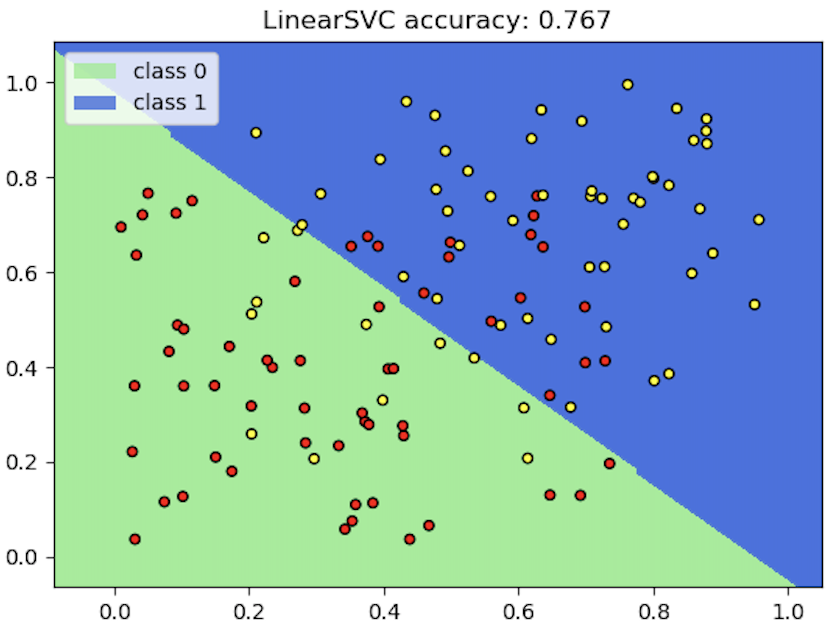} \\
(c) SCD01 complex & (d) SVM complex \\
\multicolumn{2}{c}{\includegraphics[width=\columnwidth]{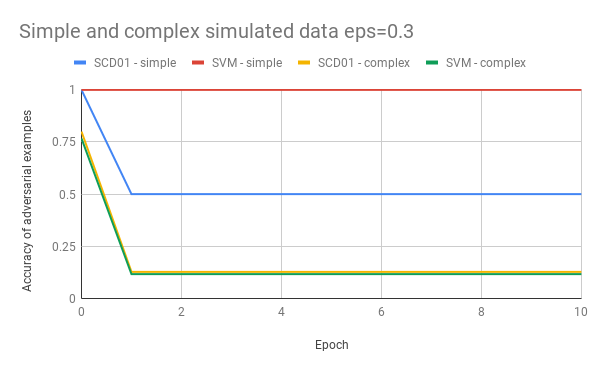}}\\
\multicolumn{2}{c}{(e)}
\end{tabular}
\vspace{-.2in}
\caption{SCD01 and linear SVM boundaries on simple separable data in (a) and (b), and the same boundaries on complex data in (c) and (d). In (e) we show the accuracy of adversarial samples generated at each epoch during substitute model training on the simple and complex datasets. As before at epoch 0 we have the accuracy of the target model on clean test data (without adversaries).}
\label{mnist2}
\end{center}
\vskip -0.2in
\end{figure}

In Figure~\ref{mnist2} we show SCD01 and SVM boundaries on simple and complex simulated datasets. On the simple dataset (shown in Figure~\ref{mnist2}(a) and (b)) we see that the SCD01 boundary is close to one class whereas the SVM is centered to maximize the margin. This is due to the non-uniqueness of the 01 loss function. There are infinite solutions on the simple dataset and the search ends as soon as the loss value becomes zero. On the complex dataset however both SCD01 and SVM boundaries are similar. 

\setlength{\tabcolsep}{1pt}

We conjecture that attacking SCD01 on a simple dataset would be easy because a convex substitute model (such as MLP and SVM) would have a boundary similar to the SVM. Thus adversaries from its boundary are likely to succeed in attacking SCD01 whose boundary lies close to one class. Indeed we see in Figure~\ref{mnist2}(e) that SCD01 falls from 100\% to 50\% accuracy on the simple dataset after the first epoch of the black box attack whereas SVM remains at 100\%. 
We now consider adversarial training to improve SCD01's robustness particularly on simple MNIST type datasets.

\subsection{Adversarial training}
\begin{figure}[h!]
\begin{center}
\begin{tabular}{c}
\includegraphics[width=\columnwidth]{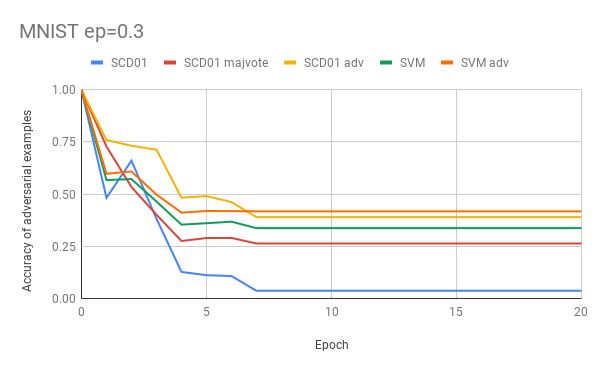} \\ 
\vspace{-.4in}
\\
\includegraphics[width=\columnwidth]{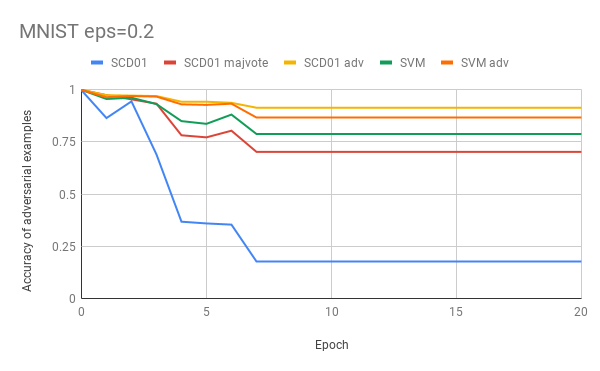} \\
\vspace{-.4in} 
\\
\includegraphics[width=\columnwidth]{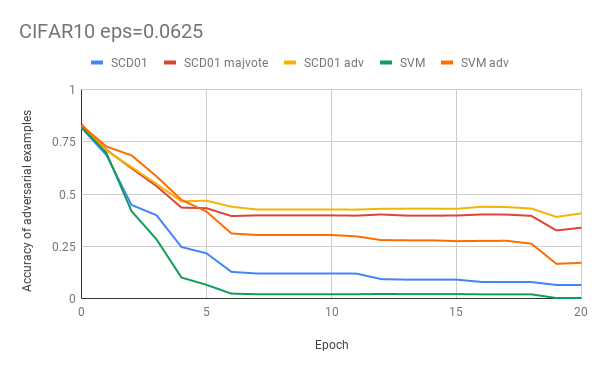} \\
\end{tabular}
\vspace{-.15in}
\caption{Accuracy of adversarial samples after iterative adversarial training on MNIST with two different distortions and CIFAR10}
\label{adv1}
\end{center}
\vskip -0.2in
\end{figure}

In order to increase the robustness of SCD01 we run Algorithm~\ref{adv} starting with a single run SCD01 trained on the full training dataset. We then take the majority vote of all 100 classifiers learnt in each iteration. We also run the same algorithm by replacing SCD01 with SVM and instead of voting across all 100 we use just the final model  as is typical in adversarial training \cite{kurakin2016adversarial}. We find the adversarially trained SCD01 is more robust than SCD01 and SCD01 majvote on all datasets while retaining clean test accuracy. In Figure~\ref{adv1} we see the adversarially trained SCD01 on MNIST and CIFAR10 outperforms the versions trained on the clean data. It is comparable to the adversarially trained SVM on MNIST  $\epsilon=0.3$ and better on MNIST $\epsilon=0.2$ and CIFAR10. 

\vspace{-.1in}
\setlength{\tabcolsep}{6pt}

\subsection{Transferability of adversaries between 01 and convex loss}

\setlength{\tabcolsep}{3pt}
\begin{table}[h!]
\caption{Transferability of black box adversaries between 01 and convex loss in CIFAR10 and STL10 (both $\epsilon=0.0625$)}
\label{transfer}
\begin{center}
\begin{small}
\begin{sc}
\begin{tabular}{cccc} \toprule
           & \multicolumn{3}{c}{CIFAR10} \\ 
Adversarial           & \multicolumn{3}{c}{Black box target model} \\
accuracy of model & MLP01 (32 votes) & MLP & SVM \\ \midrule
MLP01 (32 vo.)   & 59.7\% & 55.8\% & 60.8\%  \\
MLP & 52\% & 8.1\% & 8.4\% \\
SVM    & 32.3\% & 1.8\% & 0.17\% \\ \midrule
           & \multicolumn{3}{c}{STL10} \\ 
Adversarial           & \multicolumn{3}{c}{Black box target model} \\
accuracy of model & MLP01 (32 votes) & MLP & SVM \\ \midrule
MLP01 (32 vo.)   & 60.5\% & 58.5\% & 64.93\%  \\
MLP & 52.9\% & 14.9\% & 13.1\% \\
SVM    & 46.6\% & 7.2\% & 2.5\% \\
\bottomrule
\end{tabular}
\end{sc}
\end{small}
\end{center}
\vskip -0.1in
\end{table}

Adversarial samples are known to transfer between classifiers \cite{papernot2016transferability}. We find this is not so true for 01 loss adversaries. In Table~\ref{transfer} we see that adversaries targeting MLP01 also attack MLP and SVM but to a lower extent than if we attacked MLP and SVM directly as the target model. Adversaries produced by attacking MLP and SVM transfer between each other but not to MLP01. 

\subsection{Runtimes, stability, and comparison to prior work}
In Table~\ref{runtimes} we see that our solver with majority vote is considerably faster than the previous one \cite{xie2019} (100 votes) but still slower than the convex counterparts. These are measured on Intel Xeon Silver 4114 CPUs with NVIDIA Titan RTX 2080 GPUs.
 
\vspace{-.2in}
\setlength{\tabcolsep}{2pt}
\begin{table}[h!]
\caption{Runtimes of prior work \cite{xie2019} denoted by PREV-SCD01 and our methods shown in hours}
\vspace{-.15in}
\label{runtimes}
\begin{center}
\begin{small}
\begin{sc}
\begin{tabular}{cccccc} \toprule
Data & PREV-SCD01 & SCD01 & MLP01 & SVM & MLP \\ 
         &  100 votes & 100 votes & 32 votes & & \\ \midrule
CIFAR10    & 13.94 & 1.74 & 2.57  & 0.003 & 0.01 \\
STL10 & 12.23 & 1.38 & 0.71 & 0.004 & 0.008 \\
MNIST   & 3.1 & 0.26 & 0.68  & 3E-4 &  0.001 \\ \bottomrule
\end{tabular}
\end{sc}
\end{small}
\end{center}
\vskip -0.1in
\end{table}

We compare the adversarial accuracy of the previous 01 loss solver to ours on CIFAR10 and STL10 and find the robustness to be similar (see Supplementary Material for graph). This further supports our hypothesis of 01 loss robustness over convex ones since we see a high robustness across two different 01 loss solvers. In Table~\ref{stddev} we see that both SCD01 and MLP01 majority vote on the test data have low deviation suggesting that our results are stable and reproducible.


\setlength{\tabcolsep}{1pt}
\begin{table}[h!]
\caption{Mean and standard deviation of 100 votes of SCD01 and 32 votes of MLP01}
\vspace{-.1in}
\label{stddev}
\begin{center}
\begin{small}
\begin{sc}
\begin{tabular}{ccccccc} \toprule
&  \multicolumn{3}{c}{SCD01 (100 votes)} & \multicolumn{3}{c}{MLP01 (32 votes)} \\
 & STL10 & CIFAR10 & MNIST & STL10 & CIFAR10 & MNIST \\ 
Mean & .75 & .81  &  .99   & .76   & .83  &  .99 \\
Std dev & .008 & .005  &  8E-3 &  .01 & .006  &  8E-3 \\ \bottomrule
\end{tabular}
\end{sc}
\end{small}
\end{center}
\vskip -0.1in
\end{table}

\subsection{Applications: street sign and facial recognition adversarial attacks}
\begin{figure}[h!]
\begin{center}
\centerline{\includegraphics[width=\columnwidth]{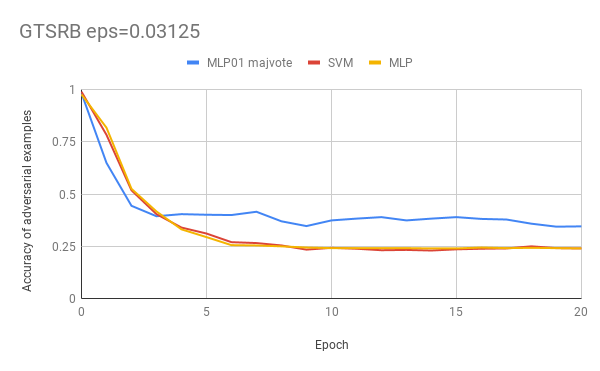}}
{\footnotesize
\begin{tabular}{ccccc}
Epoch & MLP01majvote & SVM & MLP \\
0 & .98 & .99 & .98  \\
20 & .35 & .24 & .24 \\
\end{tabular}
}
\centerline{\includegraphics[width=\columnwidth]{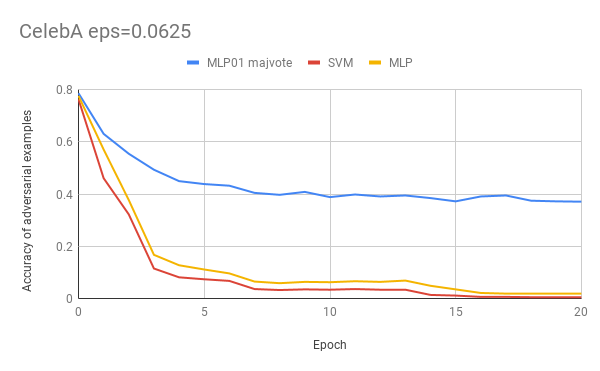}}
{\footnotesize
\begin{tabular}{ccccc}
Epoch & MLP01majvote & SVM & MLP \\
0 & .79 & .76 & .78 \\
20 & .37 & 0 & .02 \\
\end{tabular}
}
\vspace{-.1in}
\caption{Accuracy of adversarial samples on GTSRB and CelebA (see Figure~\ref{alldata} caption for more)}
\label{applications}
\end{center}
\vskip -0.2in
\end{figure}

\begin{figure}[h!]
\begin{center}
\begin{tabular}{cccc}
Clean & MLP01 majvote & MLP & SVM \\
\includegraphics[scale=.55]{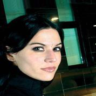} & \includegraphics[scale=.55]{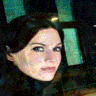} & \includegraphics[scale=.55]{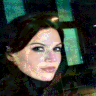} & \includegraphics[scale=.55]{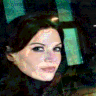} \\
\includegraphics[scale=1.2]{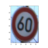} & \includegraphics[scale=1.2]{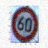} & \includegraphics[scale=1.2]{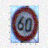} & \includegraphics[scale=1.2]{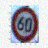} \\
\end{tabular}
\vspace{-.1in}
\caption{Images from CelebA in the top row and GTSRB in the bottom one (see Figure~\ref{sampleimages} caption for more)
}
\label{sampleimages2}
\end{center}
\vskip -0.2in
\end{figure}

We now turn to two practical problems where adversarial attacks pose a problem. First is the task of street sign detection by autonomous vehicles \cite{sitawarin2018darts} and the second is facial recognition that are used by government and security systems. We consider 2816 train and 900 test $48\times48\times3$ images of street signs of 60 and 120 mph from the GTSRB street sign dataset \cite{Stallkamp-IJCNN-2011} and 1000 train and 1000 test $96\times96\times3$ images of brown and black hair individuals from the CelebA facial recognition benchmark \cite{liu2015faceattributes}. For GTSRB we use a perturbation of $\epsilon=0.03125$ and for CelebA we use $\epsilon=0.0625$ in the black box attack. We show other values of $\epsilon$ in the Supplementary Material and make similar observations as here. 
In Figure~\ref{applications} we see that the MLP01 attains comparable accuracy to SVM and MLP but is more robust as we saw in earlier benchmarks. We show sample adversarial images in Figure~\ref{sampleimages2}.

\setlength{\tabcolsep}{6pt}

\setlength{\tabcolsep}{1pt}

\section{Discussion}
Our results show that a convex substitute model (like the multilayer perceptron that we use) can generate effective adversaries for other convex ones like SVM and MLP but not so much on 01 loss like our SCD01 and MLP01. We ask two  follow-up questions. (1) Can we attack the 01 loss with a 01 loss substitute model? (2) Was the multilayer perceptron substitute model in the black box attack correctly trained? To answer the first question we use SCD01 as the substitute model in our black box attack. 

\setlength{\tabcolsep}{1pt}
\begin{figure}[h!]
\begin{center}
\begin{tabular}{c}
\includegraphics[width=\columnwidth]{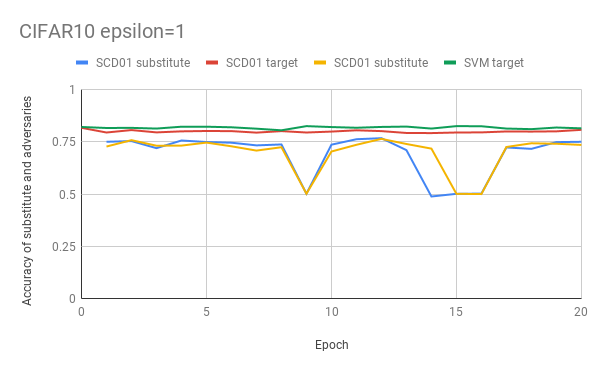} \\
\end{tabular}
\vspace{-.2in}
\caption{Accuracy of adversarial samples generated at each epoch during SCD01 substitute model training on CIFAR10 with distortion $\epsilon=1$. At epoch 0 we have the accuracy of the target model on clean test data (without adversaries). Also shown is the accuracy of the SCD01 substitute model on test samples during training to confirm that the substitute model has test accuracy comparable to the target model.}
\label{scdself}
\end{center}
\vskip -0.2in
\end{figure}

In Figure~\ref{scdself} we see the results of attacking SCD01 and SVM with SCD01 as the substitute model in the black box attack. We use the same seed for the random number generator in SCD01 both for the target and substitute model to avoid differences due to randomness. We don't use the majority vote and attack a single run of SCD01. 

We see that the black box attack with SCD01 as the substitute model fails to attack any of the models even though the accuracy of the substitute on test data is high during training and the distortion is set to a high value of $\epsilon=1$. We argue this is because of its non-unique nature: there can be infinite solutions all yielding the same local minimum in the 01 loss search space. Thus when we attempt to learn an SCD01 single vote model to generate adversaries we find it cannot even approximate and successfully attack the same SCD01 trained on clean data with the same random number generator seed as the substitute.

Of course we can attack SCD01 if we know its model parameters $(w,w_0)$. We simply generate adversaries with $x_i' = x_i + (-y'_i)\frac{w}{||w||}$ and these will fool the SCD01 single vote. In Table~\ref{whitebox} we see that the SCD01 adversaries generated in this way fool the SCD01 classifier and also transfer over to the SVM to some degree. 
This would be a white box attack though. If the model parameters are kept hidden or retrained we see that both convex and 01 loss substitute models find it hard to attack 01 loss. 

\vspace{-.1in}
\setlength{\tabcolsep}{2pt}
\begin{table}[h!]
\caption{Accuracy of adversarial examples produced by the SCD01 $w$ vector in CIFAR10 (white box attack). In parenthesis are test data accuracies without adversaries.}
\label{whitebox}
\vspace{-.1in}
\begin{center}
\begin{small}
\begin{sc}
\begin{tabular}{ccccc} \toprule
Attack model & SCD01 & SVM & MLP01  & MLP  \\ 
SCD01 (white box) & .19 (.83) & .79 (.82) & .86 (.86) & .87 (.89) \\ 
\bottomrule
\end{tabular}
\end{sc}
\end{small}
\end{center}
\vskip -0.1in
\end{table}

\begin{figure}[ht]
\begin{center}
\includegraphics[trim=0 20 0 0, clip, width=\columnwidth]{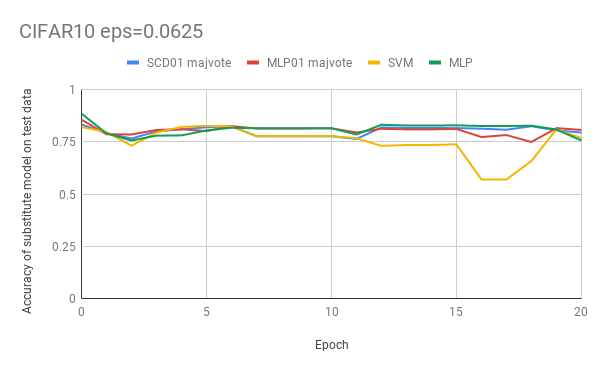} 
\vspace{-.2in}
\caption{Accuracy of the substitute model on CIFAR10 while attacking SCD01, MLP01, SVM, and MLP}
\label{subst}
\end{center}
\vskip -0.2in
\end{figure}

For the second question we look at  the accuracy of our multilayer perceptron substitute model in each epoch of the black box attack. An accurately trained substitute model indicates that our training was successful and its gradient is likely to be an effective generator of adversaries. Indeed we see in Figure~\ref{subst} that the black box substitute model accuracy while attacking SCD01, MLP01, SVM, and MLP on CIFAR10 are similar to the clean test accuracies of the target model suggesting we have a well-trained substitute. We see the same trend on all datasets.

\setlength{\tabcolsep}{1pt}

We have shown that substitute model black box attacks are not so effective against 01 loss models when the substitute model is convex or 01 loss. There are however other black box attack methods that rely on just labels and try to estimate the minimum distortion of adversarial examples (an NP-hard problem) \cite{chen2017zoo,brendel2017decision,chen2019hopskipjumpattack}. We obtained the implementations of the Boundary Attack \cite{brendel2017decision} and HopSkipJump attack \cite{chen2019hopskipjumpattack} to determine the minimum adversarial distortion of our SCD01 boundary. Both of these start with an adversarial example and make incremental changes until the example is just at the boundary of the target model. In our initial attempts we found both codes to crash before reaching convergence or their default maximum iterations when we attack our SCD01 and SCD01 majvote models. While we need to revisit the attacks both are slow even for a single example. This is not surprising since finding the minimum distortion is an NP-hard problem and thus hard to solve in practice.


We measure the $L_2$ distances between adversarial and clean images shown in Figure~\ref{sampleimages} plus the Celeba images in Figure~\ref{sampleimages2} and average them: MLP01=11.54, MLP=11.5, SVM=11.5. Thus despite MLP01 adversaries have a higher distortion they are still correctly classified by MLP01.

We have not shown the effect of different parameters on SCD01 and MLP01 because our focus here is on adversarial attacks on 01 loss. We determined our parameters by optimizing accuracy on the test dataset and then fix them for the adversarial attacks. While our SCD01 and MLP01 are possibly the fastest 01 solvers that we know of, our runtimes are still higher than SVM and MLP. Thus speeding up our algorithms by parallelization is a key part of future work.

There are several other future avenues we could explore going forward. The first is multi-class classification so we can evaluate 01 loss on full image benchmarks. In previous work \cite{xie2019} a one-vs-one strategy of 10 votes showed promising but limited results. We instead plan to add more nodes to the final layer of MLP01 and rely on majority vote for classification. Even though our results here are for classes 0 and 1 we expect similar trends on other pairs of classes in the benchmarks. Besides a multi-class network we may explore 01 loss convolutions in an attempt to match the accuracy of convolutional neural networks. This is computationally significantly hard but perhaps possible by extending the stochastic coordinate descent like we have in this study.

We briefly touch upon adversarial training in this paper and plan to explore it separately. In particular we plan to study adversarially trained SVM and MLP, and explore iterative training for SCD01 more thoroughly. It is unclear how to generate white box adversaries for MLP01 and so a naive iterative training like we did for SCD01 will not work there. One strategy is to use gradient free black box attacks but runtime may be a problem. If we can successfully adversarially train MLP01 it may become more robust than what we demonstrate here.




\newpage
\bibliography{my_bib}
\bibliographystyle{icml2020}


\end{document}